\documentclass[12pt,letterpaper]{article}

\usepackage{arxiv}

\usepackage[utf8]{inputenc} 
\usepackage[T1]{fontenc}    
\usepackage{url}            
\usepackage{booktabs}       
\usepackage{amsmath,amsfonts,amssymb}       
\usepackage{nicefrac}       
\usepackage{microtype}      
\usepackage{lipsum}

\usepackage{times}
\usepackage{epsfig}
\usepackage{graphicx}
\usepackage{amsmath}
\usepackage{amssymb}
\usepackage{comment}
\usepackage{multirow}

\usepackage{float}
\usepackage{bm}
\usepackage{array}

\usepackage{booktabs}
\usepackage{caption}
\usepackage[inline]{enumitem}

\usepackage[pagebackref=true,breaklinks=true,letterpaper=true,colorlinks,bookmarks=false]{hyperref}
\usepackage[nameinlink]{cleveref}

\begin{document}

\title{Pedestrian Intention Prediction: A Multi-task Perspective}

\author{
  Smail Ait Bouhsain\thanks{Equal contribution} \\ EPFL, VITA\\ \texttt{smail.aitbouhsain@alumni.epfl.ch} \\ \And
  Saeed Saadatnejad\textsuperscript{*} \\ EPFL, VITA\\ \texttt{saeed.saadatnejad@epfl.ch} \\ \And
  Alexandre Alahi \\ EPFL, VITA\\ \texttt{alexandre.alahi@epfl.ch} \\
}

\maketitle

\begin{abstract}

In order to be globally deployed, autonomous cars must guarantee the safety of pedestrians. This is the reason why forecasting pedestrians' intentions sufficiently in advance is one of the most critical and challenging tasks for autonomous vehicles.
This work tries to solve this problem by jointly predicting the intention and visual states of pedestrians.
In terms of visual states, whereas previous work focused on x-y coordinates, we will also predict the size and indeed the whole bounding box of the pedestrian.
The method is a recurrent neural network in a multi-task learning approach. It has one head that predicts the intention of the pedestrian for each one of its future position and another one predicting the visual states of the pedestrian.
Experiments on the JAAD dataset show the superiority of the performance of our method compared to previous works for intention prediction.
Also, although its simple architecture (more than 2 times faster), the performance of the bounding box prediction is comparable to the ones yielded by much more complex architectures.
Our code is available online \footnote{\url{https://github.com/vita-epfl/bounding-box-prediction}}.

\end{abstract}


\section{Introduction}

Future prediction is usually considered as an essential part of intelligence \cite{bubic2010prediction}.
It becomes more critical in autonomous cars as it can avoid accidents with humans.
For example, consider a situation when a pedestrian is next to a street and is going to cross.
A non-predictor agent might recognize the pedestrian just when he/she is in front of it and then tries to avoid the collision.
However, when a predictor agent looks at the same scene, it goes beyond the pedestrian detection and predicts what happens in the next few seconds. Therefore, it finds out the intention of that person and then, based on that, decides when to stop and when to pass.
In the application of autonomous cars, there are different levels of pedestrian prediction, such as its intention and visual states.

\textbf{Pedestrian intention prediction.}
To guarantee the safety of pedestrians, a self-driving car should not only predict their intention but also predict it sufficiently in advance in order to react accordingly.
In most papers of this category \cite{evans2003predicting, neogi2017context, fang2018pedestrian, hoy2018learning}, either they do intention estimation, or the prediction is performed only for a short horizon as they are considering the current intention. However, we tackle this problem by providing a sequence of predictions of the intention for the next few frames. This leads to a more extended and more accurate prediction.

\textbf{Visual states prediction.}
There are different levels of state prediction.
The problem of trajectory prediction is defined as forecasting a sequence of future positions (x-y coordinates) of a pedestrian given a set of observed past positions.
It is not a trivial task for autonomous systems because of the different kinds of paths that humans choose with non-linear speed variation. Numerous methods have been proposed.
Some are model-based \cite{helbing1995social,pellegrini2009you}
which are scenario-specific and have low performance in approximating complex functions such as pedestrian trajectory.
Other methods are data-driven, using Long Short Term Memory (LSTM) \cite{alahi2016social, xu2018encoding} or convolutional neural networks (CNN) \cite{nikhil2018convolutional}, or both \cite{xue2018ss, manh2018scene, wang2017trajectory}. These models usually have higher performance in discovering complex patterns, and we follow this approach.
Moreover, some works go beyond the trajectory and predict not only future locations (x,y) but also the width and height (w,h). We refer to this task as the bounding box prediction \cite{bhattacharyya2018long, styles2020mof}. The accuracy in this task is lower than the trajectory prediction task as this is a harder task.
We argue that the ability of visual states prediction is useful in better understanding of pedestrian intention.

In this paper, we propose a data-driven method based on a multi-task learning \cite{caruana1997multitask} approach to do efficiently pedestrian intention prediction and pedestrian bounding box prediction simultaneously.
We will show how those learned representations share some useful information to enhance the performance of each other.
For instance, if we know that a person is going to cross the road, it will be easier to predict its trajectory and vice versa.
The network is carefully designed such that it has much fewer parameters while maintaining a higher or comparable accuracy over the state-of-the-art.
\label{sec:intro}
\section{Related Work}

Most previous works in pedestrian prediction \cite{alahi2016social, xue2018ss, manh2018scene} considered pedestrians seen from a bird view. However, in this paper, we predict positions seen from a car camera view. This allows for a more complex representation of pedestrians who can be represented by bounding boxes. Indeed, bounding box center, width, and height can provide information on the position, the orientation of the pedestrian, and its distance from the car. This is valuable for a number of applications and can improve the prediction of pedestrians' intentions as well. This perspective, however, adds a challenge of car movement, making the position of the bounding box vary even if the pedestrian stops.

\subsection{Pedestrian Intention Prediction}
There are model-based approaches using social behavior analysis \cite{evans2003predicting}, social force modeling \cite{helbing1995social} and conditional random fields \cite{neogi2017context} for pedestrian dynamic prediction.
These models are scenario-specific and have low performance in approximating complex functions.

SKLT \cite{fang2018pedestrian} is taking advantage of extracted skeleton features of pedestrians as input to a Random Forest. \cite{fang2018pedestrian} proposed a method (CNN(fc6)) using features extracted by a Faster R-CNN, as well as a combination of the two methods (CNN(fc6)+SKLT).
ConvNet-Softmax \cite{saleh2017early} used a simple CNN with softmax activation layer for classification. \cite{rasouli2017they} proposed the same method with an SVM classifier instead of the softmax layer (ConvNet-SVM). \cite{carreira2017quo} replaced the SVM with an LSTM (ConvNet-LSTM) and ST-Dense-Net \cite{saleh2019real} combined the spatial and temporal features of the observed sequence.
In this paper, we show that our method performs better than all these models.

There are also some other works that we could not compare with because of the different setups \cite{kwak2017pedestrian, hoy2018learning}.
\cite{kwak2017pedestrian} used dynamic fuzzy automata and low-level features with boosted random forests. \cite{hoy2018learning} proposed a variant of a variational recurrent neural network (VRNN), incorporating latent variable corresponding to a dynamic state space model.
A subtle calculation shows the superiority of our performance over them.

\subsection{Pedestrian Trajectory Prediction}

In trajectory prediction, the input sequence is the set of past observed positions, and the output sequence is a set of predicted positions in the future. Some previous works \cite{alahi2016social, xue2018ss, xu2018encoding} take advantage of the social neighborhood in order to improve the accuracy of the predictions, while others \cite{xue2018ss, manh2018scene, wang2017trajectory} use the scene features and layout as a predictor as well.
While most of these methods use a recurrent neural network, \cite{nikhil2018convolutional} proposed a CNN based human trajectory prediction approach. 
The go-to method \cite{alahi2016social} proposed a sequence to sequence LSTM model that combines the observed past positions and the social interactions and conventions involved in humans motion in order to predict their trajectory in crowded scenes.

\subsection{Pedestrian Bounding Box Prediction}

There is a couple of new research performing bounding box prediction. \cite{styles2020mof} introduces a method (STED) for multiple object bounding box forecasting. It is also a sequence to sequence architecture, combining both observed bounding box coordinates and optical flows in order to predict the future bounding boxes. One disadvantage this network has is the need for optical flows, which are extracted using another neural network called Flownet2 \cite{ilg2017flownet}.
\cite{styles2020mof} used two baselines: a dummy Constant-Velocity-Constant-Scale (CV-CS) model always predicting the same observed velocity and a Linear Kalman Filter (LKF) \cite{kalman1960new} applied to the bounding box prediction problem.
It also proposed adaptations of the methods proposed in \cite{styles2019forecasting} (DTP-MOF), which used optical flow as input to a CNN for bounding box prediction, and in \cite{yagi2018future} (FLP-MOF) which combined ego-motion estimation (FlowNet2) and pedestrian pose (OpenPose) as input.

In \cite{bhattacharyya2018long}, they took advantage of odometry data for predicting the bounding box of pedestrians under uncertainty. On the other hand, \cite{yao2019egocentric} used optical flow and future ego-motion of the viewer car for predicting future car bounding boxes. In both of those works, information about the motion of the car is needed, which is not always available.
\label{sec:rel}
\section{Method}

\begin{figure}[t]
    \centering
    \includegraphics[width=1.\textwidth]{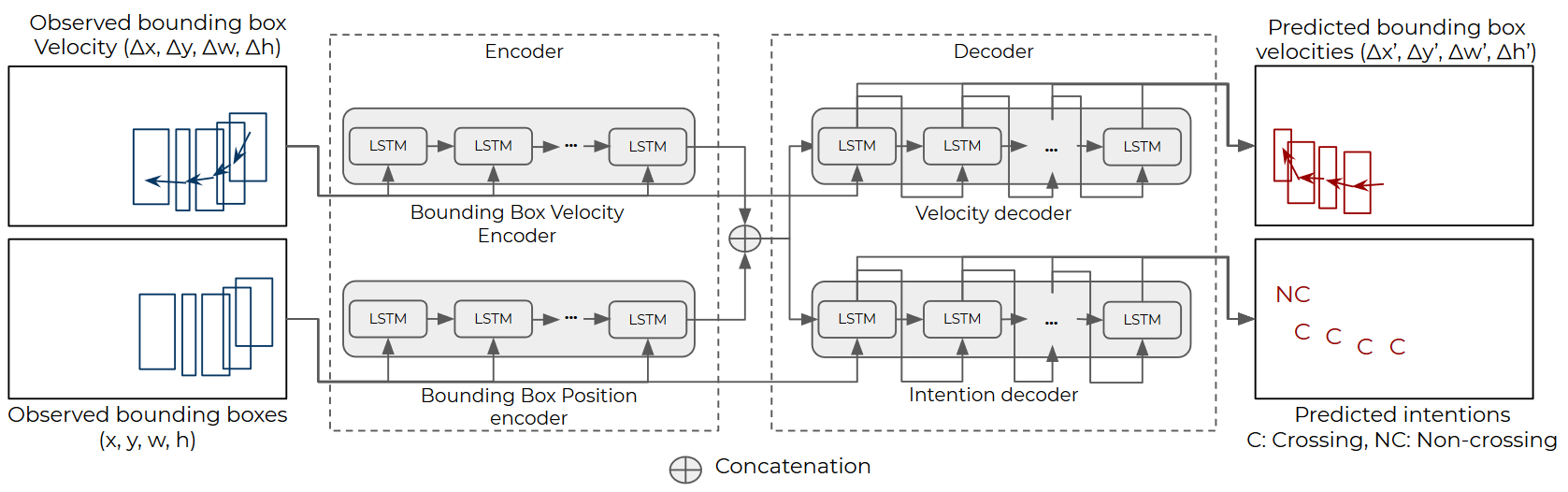}
    \caption{Our proposed multi-task learning network. It is an LSTM based encoder-decoder architecture that leverages the position and dimensions of the observed bounding box as well as their velocity in order to predict the future bounding boxes of pedestrians as well as a sequence of future intentions.}
    \label{fig:net}
\end{figure}

The proposed method (\textbf{PV-LSTM}) is a multi-task sequence to sequence LSTM model. It takes as input the velocities and coordinates of observed past bounding boxes and outputs the predicted velocities of the future bounding boxes of the pedestrian, from which the future bounding box coordinates can be computed, as well as the pedestrian's state (crossing/not crossing) in each predicted bounding boxes. A sketch of PV-LSTM is shown in Figure \ref{fig:net}, which stands for Position-Velocity-LSTM, as it encodes the position and the velocity of the pedestrian bounding box.

\subsection{Problem formulation}
Given as sequence $(B_{t-{T_{obs}}}, ..., B_{t})$ of bounding boxes of a pedestrian corresponding to the timesteps or frames $(t-{T_{obs}}, ..., t)$ in a video sequence, the task of bounding box prediction is to output the next sequence of bounding boxes $(B_{t+1}, ..., B_{t+{T_{pred}}})$ at the following timesteps $(t+1, ..., t+{T_{pred}})$, as well as the sequence of the next intentions of the pedestrian $(I_{t+1}, ..., I_{t+{T_{pred}}})$. 

The position of the bounding box of a pedestrian at timestep $t$ is represented by the coordinates of its center, its width and height $B_t = (x_t, y_t, W_t, H_t)$. The intention $I_t$ is a binary value representing the state of the pedestrian, either crossing or non-crossing, at each predicted timestep.

\subsection{Bounding box velocity encoder}
Both the position and the velocity of the pedestrian are taken into account to capture the high non-linearity in humans motion. Also, in order to tackle the car movement issue, the relative velocity of the pedestrian from the car perspective is considered.
In this paper we use LSTMs \cite{hochreiter1997long} given its power in dealing with time sequence data and long term dependencies have been proven.

The bounding box velocity LSTM encodes the velocity of the observed bounding boxes.
The latter is represented as the variation between the center positions of successive pairs of bounding boxes and height and width variations $v_{B_t} = (\Delta x_t, \Delta y_t, \Delta W_t, \Delta H_t) = (x_t - x_{t-1}, y_t - y_{t-1}, W_t - W_{t-1}, H_t - H_{t-1})$. 
Given these observed velocities as input, the velocity encoder computes the hidden state of velocity $v_t$ using:
\begin{equation}
    v_t = LSTM^{enc}_v(v_{t-1}, v_{B_t}, \bm{W}_{v}),
    \label{lstm1}
\end{equation}
where $LSTM^{enc}_v$ is the velocity encoder LSTM, and $\bm{W}_{v}$ is its weight matrix.

\subsection{Bounding box position encoder}
The position LSTM encodes the positions and dimensions of the bounding boxes throughout the input sequence. It takes as input the observed past bounding boxes coordinates $(B_{t-{T_{obs}}}, ..., B_t)$, and outputs the updated position hidden state at time $t$ via: 
\begin{equation}
    p_t = LSTM^{enc}_p(p_{t-1}, B_t, \bm{W}_{p}),
    \label{lstm2}
\end{equation}

where $LSTM^{enc}_p$ is the position encoder LSTM, and $\bm{W}_{p}$ is its weight matrix.

The position encoder is used to extract features and identify movement patterns from the input sequence of positions. These features help with the bounding box prediction but are vital components for intention prediction since the position and orientation of the pedestrian in the scene determine its state and future intentions. 

\subsection{Velocity decoder}
First, the position hidden state $p_t$ and velocity hidden state $v_t$ are concatenated resulting in one hidden state $h_t$ grouping all the features: $$h_t = p_t \oplus v_t$$
To predict the next sequence bounding box velocities of a pedestrian, $h_t$ is given as an initial hidden state to the velocity decoder $LSTM^{dec}_v$ which takes as input the last observed velocity $v_{B_t}$ and outputs the next predicted velocity of the bounding box $\hat{v}_{B_{t+1}} = (\hat{\Delta x}_{t+1}, \hat{\Delta y}_{t+1}, \hat{\Delta W}_{t+1}, \hat{\Delta H}_{t+1})$. The equations for the first prediction is computed using:
\begin{equation}
    \hat{h}^{v}_{t+1} = LSTM^{dec}_v(h_{t}, v_{B_t}, \bm{W}_{dv}),
    \label{lstm3}
\end{equation}
Then, this predicted hidden state is fed to a fully connected layer to get the output velocity via:
\begin{equation}
    \hat{v}_{t+1} = \bm{W_{ov}}\hat{h}^{v}_{t+1} + b_{ov},
    \label{yt1}
\end{equation}
where $\bm{W}_{dv}$ is the weight matrix of the the velocity decoder LSTM, $\bm{W_{ov}}$ is the weight matrix of the output layer and $b_{ov}$ its bias vector.

The next velocities are then computed iteratively while updating the hidden state, giving each time the last predicted one as input to the decoder. The hidden state is updated at time $t+t'$ using: 
\begin{equation}
    \hat{h}^{v}_{t+t'} = LSTM^{dec}_s(\hat{h}^{v}_{t+t'-1}, \hat{v}_{B_{t+t'-1}}, \bm{W}_{ds}),
    \label{lstm4}
\end{equation}
Then the output velocities are computed via:
\begin{equation}
    \hat{v}_{B_{t+t'}} = \bm{W_{ov}}\hat{h}^{v}_{t+t'} + b_{ov},
    \label{yt2}
\end{equation}

These velocities are used to compute the future position and dimension of the pedestrian's bounding box, by using cumulative addition as described in the equation below:
\begin{equation}
    \hat{B}_{t+t'} = \hat{B}_{t+t'-1} + \hat{v}_{B_{t+t'}},
    \label{yt3}
\end{equation}

\subsection{Intention decoder}
Similar to velocity decoder, this LSTM takes as initial hidden state the result of the concatenation of the encoders hidden states $h_t$. However, the first intention prediction uses as input the last observed position of the bounding box $B_t = (x_t, y_t, w_t, h_t)$, and outputs the next predicted state of the pedestrian $I_{t+1}$ via:
\begin{equation}
    \hat{h}^{I}_{t+1} = LSTM^{dec}_I(h_{t}, B_t, \bm{W}_{di}),
    \label{lstm5}
\end{equation}
\begin{equation}
    \hat{I}_{t+1} = \bm{W_{oi}}\hat{h}^{I}_{t+1} + b_{oi},
    \label{yt4}
\end{equation}
where $LSTM^{dec}_I$ is the the intention decoder LSTM, $\bm{W}_{di}$ is its weight matrix, $\bm{W_{oi}}$ is the weight matrix of the output layer and $b_{oi}$ its bias vector.

The next intentions at the next timesteps are then predicted iteratively while updating the hidden state. The last predicted state is fed to an embedding fully connected layer before being given as input to the decoder. The hidden state is updated at time $t+t'$ using: 
\begin{equation}
    \hat{E}_{t+t'-1} = \bm{W_{embedding}}\hat{I}_{t+t'-1} + b_{embedding}, 
    \label{lstm6}
\end{equation}
\begin{equation}
    \hat{h}^{I}_{t+t'} = LSTM^{dec}_s(\hat{h}^{I}_{t+t'-1}, \hat{E}_{t+t'-1}, \bm{W}_{di}),
    \label{lstm7}
\end{equation}
Then the output intentions are computed via:
\begin{equation}
    \hat{I}_{t+t'} = \bm{W_{oi}}\hat{h}^{I}_{t+t'} + b_{oi}.
    \label{yt5}
\end{equation}
where $\bm{W_{embedding}}$ is the weight matrix of the embedding layer and $b_{embedding}$ is the bias vector.

The output intentions are then fed to a softmax activation layer in order to compute the probability of each possible outcome. 

\subsection{Implementation details}
The dimension of the hidden states of LSTMs is $256$.
Our network is implemented in Pytorch and trained using the mean square error loss for bounding box prediction and the binary cross-entropy loss for intention prediction. We leverage the Adam optimizer starting at a learning rate of $0.0001$, updated by an adaptive scheduler. All the models are trained for $100$ epochs on one NVIDIA GTX-1080-Ti GPU.
\label{sec:method}
\section{Experiments}
\subsection{Datasets}
The Joint Attention in Autonomous Driving (JAAD) dataset \cite{kotseruba2016joint} consists of $346$ high-resolution videos $(1920 \times 1080)$ taken from a car camera perspective. These video clips show various typical driving scenarios in urban areas. The dataset has bounding box and action annotations for all pedestrians in the video sequences.
These videos are split into shorter ones of fixed length for each pedestrian and then divided into observed data and future data. Sequences generated from the first $300$ videos ($20000$ sequences) are our training set, and the ones from the last $46$ ($8000$ sequences) are our testing set.
This ensures that testing and training scenes do not have any intersection.

We also use the Citywalks dataset \cite{styles2020mof} in order to compare our performance in pedestrian bounding box prediction with theirs. It is not made for autonomous driving purposes. We use it mainly for comparison purposes. It contains first-person perspective videos showing pedestrians in various public areas. The dataset has been annotated using Mask-RCNN network for bounding box tracking.

\subsection{Baselines and Evaluation Metrics}
\label{section:baselines}
We compare the performance of our model to the following baselines as well as other published works used for pedestrian bounding box and intention prediction:
\begin{itemize}

  \item \textbf{Pedestrian intention prediction}
  We use the results reported in \cite{saleh2019real} in order to evaluate the performance of our model for pedestrian intention prediction. We also compare our performance with \cite{fang2018pedestrian}.
  
  We add a baseline (\textbf{Trajectory-LSTM}) in which the bounding box prediction ($x, y, w, h$) is replaced by trajectory prediction ($x, y$) in order to verify the effectiveness of our approach.

  \item \textbf{Pedestrian bounding box prediction:} 
  \begin{itemize}
      \item \textbf{Evaluation on JAAD:} For comparison with the following baselines, the models take as input $18$ observations ($0.6$ seconds) and outputs $18$ predictions ($0.6$ seconds) :
      \begin{itemize}
          \item \textbf{Position-LSTM:} This model predicts the position of future bounding boxes directly given the observed past ones.
        
          \item \textbf{Velocity-LSTM:} This network takes as input the velocities of bounding boxes only and outputs the predicted future velocities. This is similar to the ablation of the position encoder and the intention decoder of our model.
          
          \item \textbf{Scene-PV-LSTM:} It adds a scene features encoder to our proposed model, by combining Resnet-50, another CNN for dimensionality reduction and an LSTM. The scenes are rendered by drawing a red bounding box around the pedestrian in question. 
     \end{itemize}
     \item \textbf{Evaluation on Citywalks:} 
     As there is no previous work doing the same experiments in JAAD, the model is evaluated on Citywalks. We test how our model deals with long term predictions and compare the performance with those reported in the recent work \cite{styles2020mof} on their proposed Citywalks dataset.
      
  \end{itemize}

\end{itemize}

We use the average displacement error (ADE) and the final displacement error (FDE) as evaluation metrics for the performance of the center prediction of the future bounding boxes. We evaluate the performance of the dimensions prediction using the average and final intersection over union (AIOU, FIOU). For the pedestrian intention prediction task, we use accuracy as the evaluation metric.

\subsection{Results}
\subsubsection{Quantitative Results}
The quantitative results of our experiments in pedestrian intention prediction can be found in Table \ref{table:1} and \ref{table:2}. The ones in pedestrian bounding box prediction are in Tables \ref{table:3} and \ref{table:4}. We evaluate the performance of each task without considering the other head (\textit{Single-task}) and with the whole network (\textit{Multi-task}). 

\begin{table}[t]
\centering
\caption{The accuracy performance of the intention prediction on JAAD dataset. Each method takes as input 14 observations and makes one future intention prediction.}
 \begin{tabular}{c c} 
 \hline
 Model & Accuracy (\%) \\
 \hline
 CNN(fc6) \cite{fang2018pedestrian} & 70.0\\
 SKLT \cite{fang2018pedestrian} & 88.0\\
 CNN(fc6)+SKLT \cite{fang2018pedestrian} & 87.0\\
 Trajectory-LSTM \ref{section:baselines}  & 86.16\\
 \hline
 Single-task PV-LSTM (ours) & 89.67\\
 Multi-task PV-LSTM (ours) & \textbf{91.48}\\
 \hline
 
\end{tabular}
\label{table:1}
\end{table}

\begin{table}[t]
\centering
\caption{The run-time performance of the network on JAAD dataset.}
 \begin{tabular}{c  c  c} 
 \hline
 Model & Runtime (ms)\\
 \hline
 ConvNet-Softmax \cite{saleh2017early} & 28ms\\
 ConvNet-SVM \cite{rasouli2017they} & 27ms\\
 ConvNet-LSTM \cite{carreira2017quo} & 40ms\\
 C3D \cite{carreira2017quo} & 27ms\\
 ST-Dense-Net \cite{saleh2019real} & 10ms\\
 Trajectory-LSTM \ref{section:baselines} & 4.9ms \\
 \hline
 Single-Task PV-LSTM (ours) & \textbf{4.1ms}\\
 Multi-Task PV-LSTM (ours) & \textbf{4.7ms}\\
 
 \hline
 
\end{tabular}
\label{table:2}
\end{table}

For comparison with all different models presented in \cite{fang2018pedestrian}, we modified the experiment setups as in \ref{table:1}. As Table \ref{table:1} shows, the accuracy of our model in intention prediction outperforms all the baselines \cite{fang2018pedestrian}.  
This is because our model is not limited to one prediction. It takes advantage of multiple predictions in the future to determine with more accuracy the immediate future intention of the pedestrian.
This suggests that gaining more accuracy is possible without needing more features such as skeleton or scene features.
As they did not provide timings, we cannot compare our run-time.

From Tables \ref{table:1} and \ref{table:2}, we conclude that bounding box prediction (\textbf{Multi-task PV-LSTM}) instead of trajectory prediction (Trajectory-LSTM) could successfully improve the intention prediction without penalizing the run-time.

\begin{table}[t]
\centering
\caption{Comparison of the performances of our models and the baselines on the JAAD dataset for the pedestrian bounding box prediction task. All the results reported are the evaluation metrics of the models on the testing set. ADE and FDE are in pixels in which the lower is better.}
\begin{tabular}{ccccc}
 \hline
 Model & ADE & FDE & AIOU (\%) & FIOU (\%) \\
 \hline
 Position-LSTM \ref{section:baselines} & 31.5 & 46.5 & 39.0 & 29.9 \\
 Velocity-LSTM \ref{section:baselines} & 9.28 & 15.4 & 74.6 & 63.0\\
 Scene-PV-LSTM \ref{section:baselines} & 9.78 & 15.9 & 74.1 & 61.89\\
 \hline
 Single-task PV-LSTM (ours) & \textbf{9.18} & 15.35 & 74.9 & 63.2\\
 Multi-task PV-LSTM (ours) & 9.19 & \textbf{15.22} & \textbf{75.2} & \textbf{63.3}\\
 \hline
\end{tabular}
\label{table:3}
\end{table}

\begin{table}[t]
\centering
\caption{Comparison of the performance of our models with the STED network as well as the proposed baselines in \cite{styles2020mof} on the Citywalks dataset. We also show our results for a shorter input and output sequence length (mentioned in the second column in seconds) on the same dataset. ADE and FDE are in pixels in which the lower is better.}
 \begin{tabular}{cccccc}
 \hline
 Model & (input, output) & ADE & FDE & AIOU (\%) & FIOU (\%) \\
 \hline 
 CV-CS \cite{styles2020mof} & (1,2) & 31.6 & 57.6 & 46.0 & 21.3 \\
 LKF \cite{kalman1960new} & (1,2) & 32.9 & 59.0 & 43.9 & 20.1 \\
 DTP-MOF \cite{styles2020mof} & (1,2) & 27.3 & 49.2 & 49.6 & 25.1 \\
 FPL-MOF \cite{styles2020mof} & (1,2) & 29.3 & 51.0 & 44.9 & 22.6 \\
 STED \cite{styles2020mof} & (1,2) & 26.0 & \textbf{46.9} & \textbf{51.8} & \textbf{27.5}\\
 \hline
 Single-task PV-LSTM (ours) & (1,2) & \textbf{25.2} & 49.9 & 40.2 & 20.3\\
 \hline \hline
 Single-task PV-LSTM (ours) & (0.3,0.6) & 10.22 & 17.17 & 73.2 & 62.8\\
 \hline
\end{tabular}

\label{table:4}
\end{table}

Table \ref{table:3} shows the short-term prediction results of our model in bounding box prediction compared to the baselines.  Our method has a high performance. It has a very low ADE of $9$ pixels in the resolution of $1920 \times 1080$, which is less than $1\%$ error. The high AIOU and FIOU performances show that the model precisely predicts not only the centers but also the whole bounding boxes.

Tables \ref{table:1} and \ref{table:3} show that multi-task learning improves the overall performance by introducing features from the predicted bounding boxes. Training both head networks simultaneously by back-propagating the weighted sum of the two losses helps both intention and bounding box prediction.

Table \ref{table:4} reports the results of bounding box prediction on the Citywalks dataset. We followed all implementation details mentioned in \cite{styles2020mof} to compare fairly.
The duration of the observation and prediction period are mentioned in the second column of the table.
\textbf{PV-LSTM} has a comparable performance on ADE and FDE, capturing the dependencies of the motion of the bounding box center. However, the AIOU and FIOU results are worse than the baselines reported in the previously mentioned work. This shows that our method struggles with the dimensions of the bounding box for longer predictions. The high performance of our model on shorter sequences of the same dataset confirms this.

\subsubsection{Qualitative Results}
\begin{figure}[t]
    \centering
    \includegraphics[width=1.\textwidth]{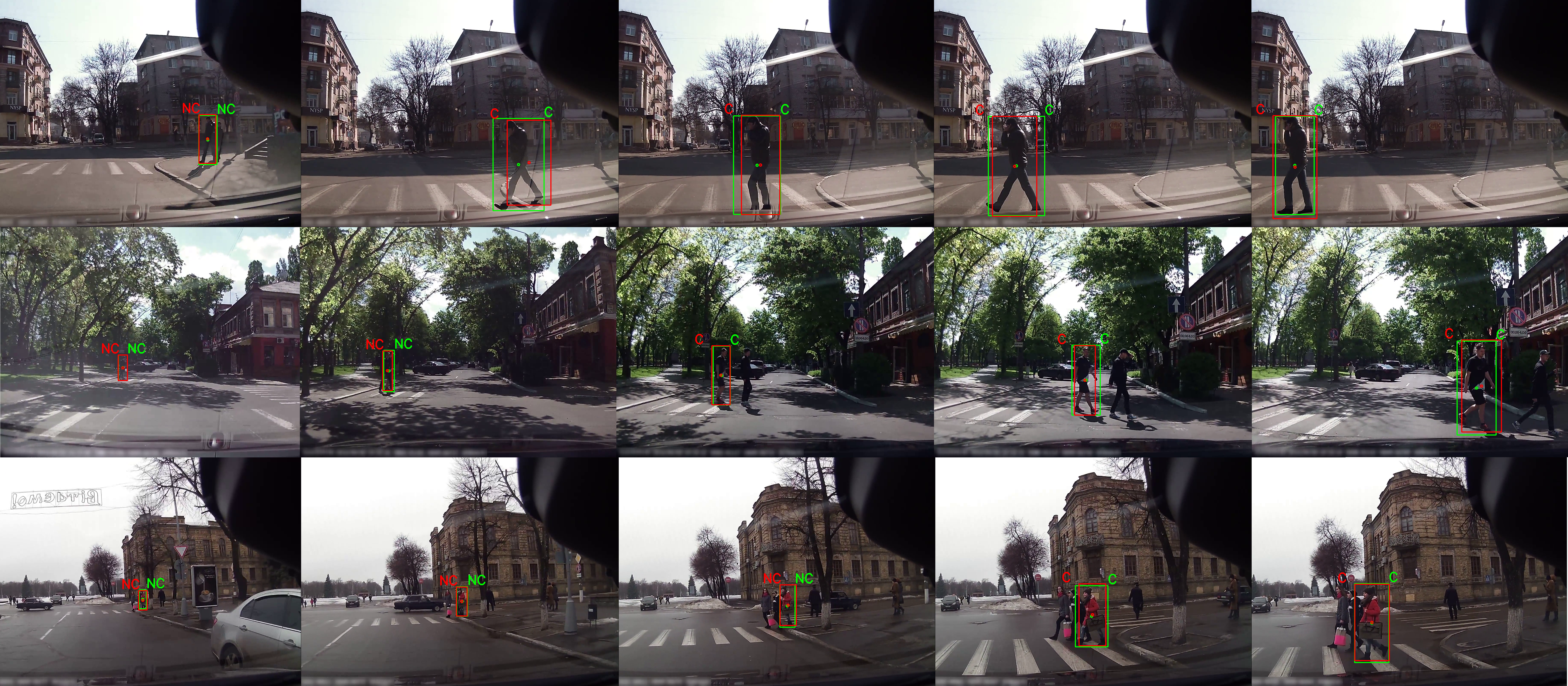}
    \caption{Visualisation of the pedestrian bounding box and intention prediction in different scenarios. Red rectangle/text: Predicted bounding box/intention | Green rectangle/text: Ground truth bounding box/intention | C : Crossing | NC: Non crossing. Each frame represents a time step, going from left to right. The first row shows how our model acts at points where the state of the pedestrian changes from crossing to non-crossing. The second shows a whole crossing sequence. The third shows how our network handles non-moving pedestrians when the car is moving.}
    \label{fig:vis}
\end{figure}

Figure \ref{fig:vis} shows some qualitative results in the dataset. The details are described in its caption. For predictions involving the pedestrians on the road, the predicted intentions are correct for all scenarios. The same thing can be observed for pedestrians on the sidewalk.

We observe that the bounding box center prediction is very accurate.
However, the dimensions prediction starts with a very high IOU, which decreases at each timestep. These illustrations are compatible with the quantitative results that we discussed.

One observed challenge in this problem is varying pedestrian bounding boxes across frames, even if the pedestrians are not moving.
Our model performs well in these cases, as shown in the last scenario of Figure \ref{fig:vis}.
It shows the ability of the model to capture car speed.
\label{sec:exp}
\section{Conclusion}
We have presented a method for performing pedestrian intention and visual states prediction in a multi-task learning approach. We have shown that our model outperforms previously proposed methods as well as state of the art in pedestrian intention prediction, and has a competitive performance for pedestrian state (bounding box) forecasting.
We demonstrated that predicting the velocity of the bounding boxes instead of the position, taking advantage of the observed positions and speed of bounding boxes, as well as using a multi-task learning architecture, could improve the accuracy of the pedestrian intention prediction.
It is also worth mentioning that our model has high accuracy while being simple with two times faster runtime.

Although our method yields better results than previous works, it still needs to be improved in longer period prediction.
In future work, we will explore how much the visual scene information can improve the predictions without adding much complexity.
Moreover, we will evaluate the proposed method on more datasets in order to validate its potential for deploying in the real-world.
\label{sec:conc}

\section*{Acknowledgements}

This project has received funding from the European Union's Horizon 2020 research and innovation program under the Marie Sklodowska-Curie grant agreement No 754354.

\bibliographystyle{unsrt}
\bibliography{references}

\end{document}